\documentclass[letterpaper]{article} % DO NOT CHANGE THIS
\usepackage[preprint]{aaai2027}  % DO NOT CHANGE THIS
\usepackage[hyphens]{url}  % DO NOT CHANGE THIS
\usepackage{graphicx} % DO NOT CHANGE THIS
\usepackage{natbib}  % DO NOT CHANGE THIS AND DO NOT ADD ANY OPTIONS TO IT
\usepackage{caption} % DO NOT CHANGE THIS AND DO NOT ADD ANY OPTIONS TO IT
\usepackage{algorithm}
\usepackage{algorithmic}

\usepackage{url}            % simple URL typesetting
\usepackage{booktabs}       % professional-quality tables
\usepackage{amsfonts}       % blackboard math symbols
\usepackage{nicefrac}       % compact symbols for 1/2, etc.
\usepackage{microtype}      % microtypography
\usepackage[table]{xcolor} 
\usepackage{multicol}
\usepackage{multirow}
\usepackage{graphicx}    % 插入图片必须
\usepackage{subfigure}
\usepackage{enumitem}
\usepackage{amsmath} 
\usepackage{newfloat}
\usepackage{listings}
\DeclareCaptionStyle{ruled}{labelfont=normalfont,labelsep=colon,strut=off} % DO NOT CHANGE THIS
\floatstyle{ruled}
\newfloat{listing}{tb}{lst}{}
\floatname{listing}{Listing}

\usepackage{booktabs}

\title{Test-Time Logit Prompting for Source-Free Missing Modality Adaptation}

\author{
Taixi Chen  \qquad
Nancy Guo\corresponding
}

\affiliations{
School of Computing\\
State University of New York at Binghamton\\
\{tchen51, nguo1\}@binghamton.edu
}

\begin{document}

\maketitle

\begin{abstract}
Vision-language models (VLMs) have achieved remarkable performance by leveraging complementary information from large-scale image-text pairs. However, missing-modality inputs are commonly encountered during real-world deployment, often leading to significant performance degradation. Existing methods primarily enhance model robustness by learning modality compensation strategies from source training data. However, their reliance on source training data makes them difficult to apply when original data are unavailable due to privacy, storage, or accessibility constraints, such as clinical applications and personalized AI services. This raises an important yet underexplored question: can VLMs be efficiently adapted at test time for visual recognition with missing modalities without accessing source training data? To this end, we propose Test-Time Logit Prompting (TLP), a lightweight source-free test-time adaptation framework for visual recognition with missing modalities. To address missing-induced prediction shifts, TLP optimizes logit prompts with uncertainty-aware adjustment and modality-complete consistency regularization, adaptively adjusting prediction confidence while preserving semantic consistency. Extensive experiments across diverse vision-language benchmarks demonstrate that TLP consistently enhances recognition performance under missing-modality scenarios, achieving up to 8\% improvements while requiring only hundreds of tunable parameters and a few test-time optimization steps.

\end{abstract}

\section{Introduction}

Vision-language models (VLMs) have demonstrated remarkable performance by leveraging complementary information from large-scale image-text pairs~\cite{clip}, achieving impressive results across various vision-language tasks, including image captioning~\cite{videocaption, videocaption1}, cross-modal retrieval~\cite{clip, vilt,  crossretreive}, and visual question answering~\cite{vqa, drivelm}. Despite these advances, real-world deployments frequently encounter missing-modality inputs caused by privacy constraints, sensor failures, or incomplete data collection~\cite{first_prompt, dcp}. The absence of certain modalities can disrupt the learned cross-modal alignment in VLMs, resulting in significant performance degradation~\cite{notrobust}. Therefore, developing robust multimodal learning frameworks that can reliably handle diverse missing-modality scenarios has become a critical challenge.

\begin{figure}[t]
    \centering
    \includegraphics[width=1\linewidth]{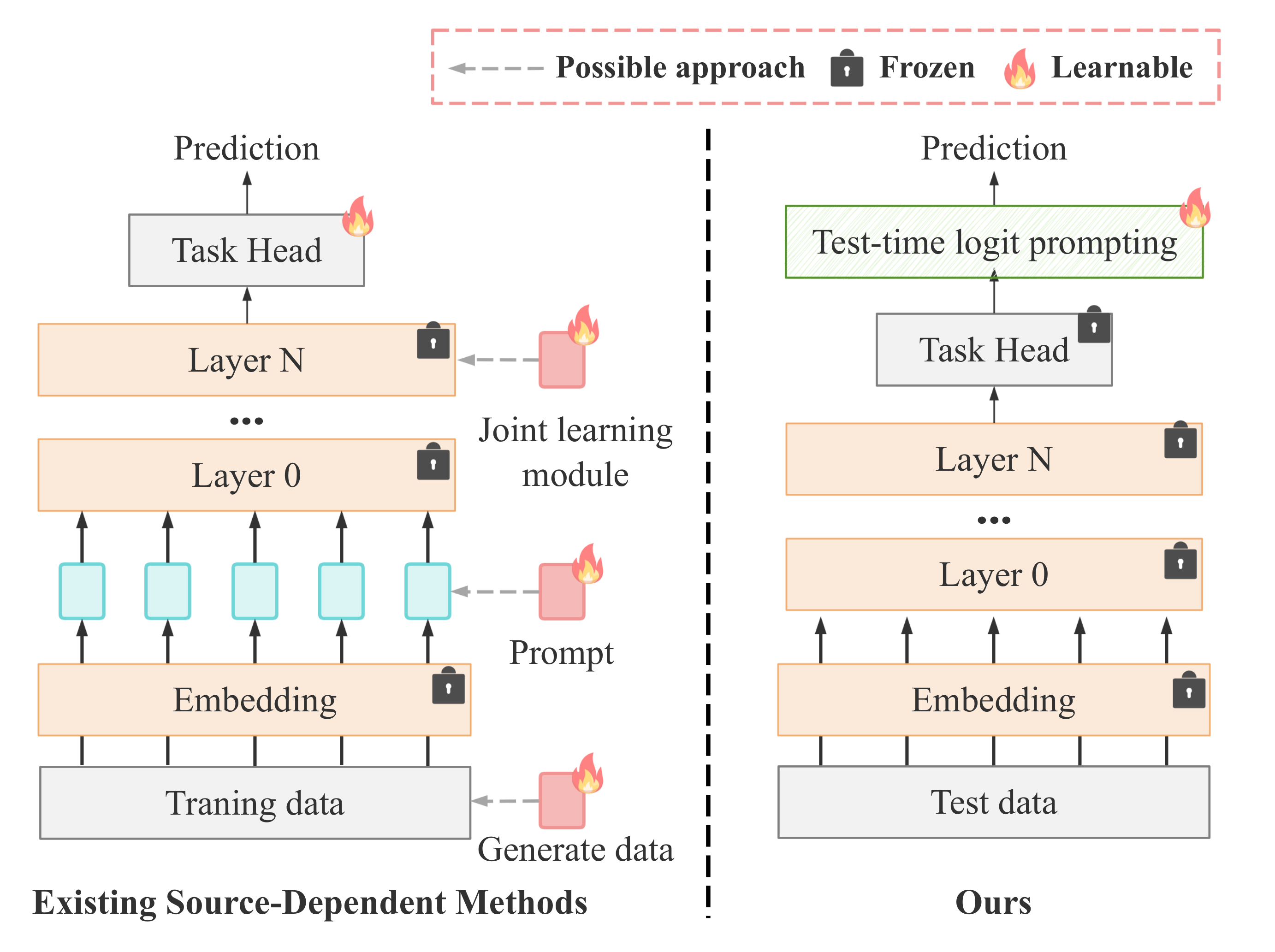}
\caption{
Comparison of the missing-modality adaptation paradigms. Existing source-dependent methods rely on training data for modality generation, prompt learning, or representation learning, whereas TLP adapts frozen VLMs at test time by optimizing lightweight logit prompts.
}
    \label{fig:compare}
    % \vspace{-0.5cm}
\end{figure}

\begin{figure*}[t]
    \centering
    \includegraphics[width=.93\linewidth]{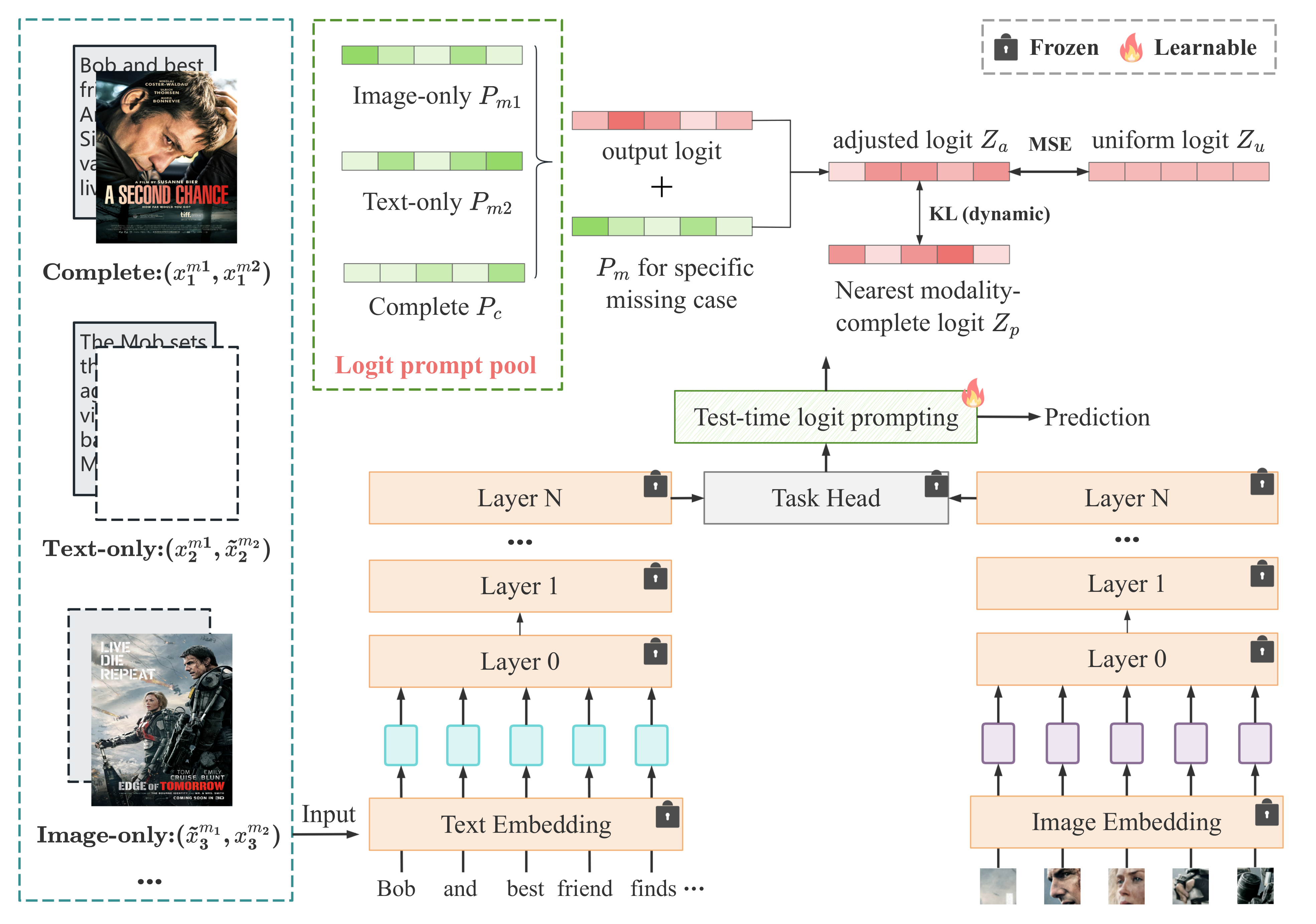}
\caption{
Overview of Test-Time Logit Prompting (TLP). Given a frozen VLM, TLP introduces missing-aware logit prompts to directly adapt the decision space under different missing-modality conditions. During inference, the corresponding logit prompt is selected according to modality availability and optimized through uncertainty-aware adjustment and modality-complete consistency regularization. The optimized prompt adjusts the original logits for final prediction, enabling efficient source-free missing-modality adaptation without updating the backbone.
}
    \label{fig:main}
    % \vspace{-0.5cm}
\end{figure*}

As illustrated in Figure~\ref{fig:compare}, existing methods mainly address this challenge through source-dependent strategies, including cross-modal generation~\cite{crossgen2, crossgen1}, joint representation learning~\cite{notrobust, representation-based, sharedfeatures}, and prompt learning~\cite{first_prompt, dcp, syp}. These approaches have effectively improved model robustness by learning or recovering missing-modality information using source training data. Despite their effectiveness, this source-dependent paradigm assumes continuous access to source training data, which is often impractical in real-world deployment scenarios due to privacy restrictions, storage limitations, or data accessibility constraints~\cite{testtime, testsurvey}, especially in privacy-sensitive applications such as healthcare systems, personalized AI services, and edge devices.

This naturally raises an important question: can VLMs further improve missing-modality robustness at test time without accessing source training data? To answer this question, we propose Test-Time Logit Prompting (TLP), a lightweight source-free adaptation framework that directly adjusts the decision space of frozen VLMs during inference. 

Specifically, instead of adapting input-level prompts or model parameters with source training data, TLP introduces lightweight missing-aware prompts in the logit space and optimizes them online with frozen VLMs to dynamically adjust predictions under different missing-modality conditions. Considering that missing modalities can introduce both confidence bias and semantic inconsistency in the prediction space, TLP balances uncertainty adjustment and modality-complete consistency regularization through efficient logit-level adaptation, enabling reliable prediction adjustment while preserving semantic information from modality-complete references. Extensive experiments on diverse vision-language benchmarks demonstrate the effectiveness and efficiency of TLP under various missing-modality scenarios. TLP consistently improves recognition performance with only hundreds of tunable parameters and a few test-time optimization steps. Moreover, it can be readily integrated with existing training-based prompt learning methods to provide further performance gains. Our contributions are summarized as follows:

\begin{itemize}
    \item We introduce a source-free test-time adaptation perspective for missing-modality recognition, enabling VLMs to improve during inference without accessing source training data.
    \item We propose Test-Time Logit Prompting (TLP), which enables decision-space adaptation by optimizing missing-aware logit prompts with uncertainty-aware adjustment and modality-complete consistency regularization.
    \item Extensive experiments demonstrate that TLP achieves consistent improvements across diverse missing-modality scenarios with only hundreds of parameters and can seamlessly enhance existing training-based methods.
\end{itemize}

\section{Related Works}

\textbf{Missing-Modality in Multi-modal Learning.}
Despite the remarkable progress in multimodal learning~\cite{multisurvey, grounded, mmrl, chen2025uam, chen2026cadm, chen2025tyrppg}, maintaining robustness under missing-modality scenarios remains challenging due to incomplete multimodal information and disrupted cross-modal alignment~\cite{multisurvey, surveyformissing, first_prompt}. Existing methods for missing-modality learning can be broadly categorized into cross-modal generation~\cite{crossgen2, crossgen1, crossgen3}, joint representation learning~\cite{notrobust, representation-based, sharedfeatures}, and prompt learning~\cite{first_prompt, context-based, dcp, syp, chen2026learning}. Cross-modal generation methods recover missing information by reconstructing missing modalities from observed ones~\cite{crossgen1, crossgen2}, with recent efforts leveraging auxiliary knowledge from complete samples for improved recovery~\cite{crossgen3}. Joint representation learning methods~\cite{representation-based, sharedfeatures} mitigate missing-modality effects by learning shared latent spaces or modality-invariant representations. Recently, prompt learning has emerged as an efficient alternative~\cite{first_prompt, syp}, adapting models through learnable prompts with limited parameter updates. Despite their effectiveness, existing methods mainly enhance robustness during training and rely on source data. In contrast, TLP explores source-free test-time adaptation to improve missing-modality recognition during inference.

\textbf{Test-Time Adaptation.}
Test-time adaptation (TTA) aims to improve model performance during inference by adapting models to test-time distributions without accessing source training data~\cite{tent}. Existing methods mainly address distribution shifts by updating normalization statistics~\cite{BN1, BN2}, optimizing model parameters~\cite{pseudo}, or adapting lightweight modules~\cite{ecotta}. However, most existing TTA methods focus on modality-complete settings, leaving test-time adaptation under missing-modality scenarios largely unexplored. Unlike prior methods that adapt internal model components, TLP performs lightweight adaptation directly in the logit space, enabling efficient missing-modality recognition with frozen VLMs.

\section{Proposed Method}

\subsection{Problem Formulation} 

We first formulate the source-free test-time adaptation setting for missing-modality recognition considered in this paper. Without loss of generality, we consider a vision-language model with two modalities, including image modality $m_1$ and text modality $m_2$. Given a source-trained multimodal model $f_{\theta}$, the original training dataset is inaccessible during adaptation, and only unlabeled test samples are available.

Under missing-modality scenarios, the test dataset can be represented as 
$\mathcal{D}_t=\{\mathcal{D}_t^c,\mathcal{D}_t^{m_1},\mathcal{D}_t^{m_2}\}$, where 
$\mathcal{D}_t^c=\{(x_i^{m_1},x_i^{m_2})\}_{i=1}^{N_c}$ denotes samples with complete modalities, while 
$\mathcal{D}_t^{m_1}=\{(x_i^{m_1})\}_{i=1}^{N_1}$ and 
$\mathcal{D}_t^{m_2}=\{(x_i^{m_2})\}_{i=1}^{N_2}$ represent samples where only modality $m_1$ or $m_2$ is available, respectively. Following previous missing-modality methods~\cite{first_prompt,dcp}, dummy inputs $\tilde{x}^{m_1}$ and $\tilde{x}^{m_2}$ (e.g., empty text or blank images) are introduced to maintain the input format of VLMs. Therefore, incomplete samples can be rewritten as 
$\tilde{\mathcal{D}}_t^{m_1}=\{(x_i^{m_1},\tilde{x}^{m_2})\}_{i=1}^{N_1}$ and 
$\tilde{\mathcal{D}}_t^{m_2}=\{(\tilde{x}^{m_1},x_i^{m_2})\}_{i=1}^{N_2}$, forming the unified test-time dataset 
$\tilde{\mathcal{D}}_t=\{\mathcal{D}_t^c,\tilde{\mathcal{D}}_t^{m_1},\tilde{\mathcal{D}}_t^{m_2}\}$.

Under this setting, the goal is to improve the performance of the source-trained model on incomplete test samples without accessing the original training data or test labels.

\subsection{Overall Pipeline}

Figure~\ref{fig:main} illustrates the overall framework of the proposed Test-Time Logit Prompting (TLP). Given a source-trained vision-language model, TLP aims to adapt model predictions under missing-modality scenarios while keeping the entire backbone frozen. Following common vision-language architectures, we adopt CLIP~\cite{clip} as the backbone, which consists of separate image and text encoders. Specifically, the input image and text are first transformed into token sequences through their corresponding pre-trained embedding layers and then processed by frozen modality-specific encoders to obtain multimodal representations.

Unlike conventional adaptation methods that update input prompts or model parameters, TLP introduces lightweight missing-aware prompts in the logit space and only optimizes these decision-space parameters during inference:
\begin{equation}
\mathcal{P}^{*}
=
\arg\min_{\mathcal{P}}
\mathcal{L}(\tilde{\mathcal{D}}_t;\theta,\mathcal{P}),
\end{equation}
where $\theta$ denotes the frozen VLM parameters, $\mathcal{P}$ represents the learnable logit prompt pool, and $\mathcal{L}$ is the source-free adaptation objective defined on unlabeled test samples. We next present the detailed design of our test-time logit prompting mechanism.

\subsection{Test-Time Logit Prompting}

Given a test sample $\tilde{x}_i$ from the unified test dataset $\tilde{\mathcal{D}}_t$, which may contain complete or dummy-filled missing modalities, the frozen VLM produces the original prediction logits:
\begin{equation}
    z_i=f_{\theta}(\tilde{x}_i),
\end{equation}
where $z_i\in\mathbb{R}^{C}$ and $C$ denotes the number of classes. Instead of adapting input representations or updating model parameters, TLP introduces lightweight missing-aware logit prompts to directly adjust the decision space. Specifically, we define a logit prompt pool $\mathcal{P}=\{P_c,P_{m_1},P_{m_2}\}$ corresponding to complete, $m_1$-available, and $m_2$-available modality conditions, respectively. Each logit prompt $P\in\mathbb{R}^{C}$ contains class-wise learnable parameters for prediction adjustment. For each test sample, the corresponding prompt $\mathbf{p}_i\in\mathcal{P}$ is selected according to its modality availability and added to the original logits:
\begin{equation}
   \hat{z}_i=z_i+\mathbf{p}_i, 
\end{equation}
where $\hat{z}_i$ represents the adjusted logits used for final prediction. During test-time adaptation, the backbone parameters $\theta$ remain frozen, and only the logit prompts in $\mathcal{P}$ are optimized.

\subsection{Source-Free Prompt Optimization}
Since source data and test labels are unavailable during adaptation, optimizing the logit prompt pool $\mathcal{P}$ requires effective self-supervised objectives. To this end, we introduce uncertainty-aware adjustment and prototype-guided regularization to adapt predictions under different missing-modality conditions.

Missing modalities can lead to unreliable predictions by causing the model to produce over-confident outputs with limited modality information. To alleviate this issue, we encourage the adjusted predictions to maintain appropriate uncertainty by reducing their deviation from a uniform distribution. Specifically, given the adjusted logits $\hat{z}_i$, the predicted probability distribution is computed as:
\begin{equation}
    q_i=\mathrm{Softmax}(\hat{z}_i).
\end{equation}
The uncertainty-aware adjustment objective is formulated as:
\begin{equation}
    \mathcal{L}_{u}
=
\frac{1}{N}
\sum_{i=1}^{N}
\frac{1}{C}
\sum_{j=1}^{C}
(q_i^{j}-u^{j})^{2},
\end{equation}
where $N$ denotes the number of test samples used for adaptation, and $\mathbf{u}\in\mathbb{R}^{C}$ denotes the uniform distribution over $C$ classes, with each element $u^{j}=\frac{1}{C}$.

However, relying only on uncertainty adjustment may overlook the semantic structure learned by the original VLM, as pushing predictions toward a uniform distribution can weaken discriminative information. To preserve semantic consistency during adaptation, we introduce modality-complete consistency regularization by leveraging complete-modality samples as reliable anchors.

For each missing-modality sample, we first identify its nearest modality-complete neighbor within the test batch according to the original prediction distribution:
\begin{equation}
    k=\arg\min_{j\in\mathcal{D}_t^c}
d(q_i,q_j^c),
\end{equation}

where $q_j^c$ denotes the prediction distribution of a complete-modality sample. The selected complete-modality prediction is used as the semantic anchor:
\begin{equation}
q_i^a=q_k^c .
\end{equation}

Then, we minimize the distribution discrepancy between the adjusted prediction and its corresponding anchor:
\begin{equation}
\mathcal{L}_{c}
=
\frac{1}{N}
\sum_{i=1}^{N}
\sum_{j=1}^{C}
(q_i^{a})^{j}
\log
\frac{(q_i^{a})^{j}}{q_i^{j}} .
\end{equation}
To balance the complementary objectives of uncertainty adjustment and modality-complete consistency, the overall optimization objective of TLP is defined as:
\begin{equation}
\mathcal{L}
=
\mathcal{L}_{u}
+
\alpha\mathcal{L}_{c}, \quad
\alpha
=
\frac{1}{C},
\end{equation}
where $C$ denotes the number of classes. The weighting factor $\alpha$ is introduced to adaptively control the contribution of the modality-complete consistency constraint. As the number of classes increases, the prediction space becomes more complex, making nearest-neighbor matching more challenging and potentially reducing the reliability of selected anchors. Therefore, setting $\alpha=\frac{1}{C}$ reduces the influence of potentially unreliable anchors while maintaining effective semantic guidance from modality-complete samples. During test-time adaptation, only the logit prompt pool $\mathcal{P}$ is optimized by minimizing $\mathcal{L}$, while all parameters of the VLM remain frozen, resulting in efficient and lightweight adaptation

% \noindent\textbf{Inference.}
After test-time optimization, the learned logit prompts are applied to adjust predictions according to the modality condition of each test sample. Specifically, the corresponding prompt $\mathbf{p}_i\in\mathcal{P}$ is selected and added to the original logits to obtain the final prediction:
\begin{equation}
    \hat{y}_i=\arg\max_{c}\mathrm{Softmax}(\hat{z}_i)^c .
\end{equation}
Since TLP only optimizes class-level logit prompts while keeping the entire VLM frozen, the number of trainable parameters is only $|\mathcal{P}|\times C$, enabling efficient adaptation with minimal computational overhead.

\section{Experiment}

\subsection{Experiment Settings}

\textbf{Datasets.}
Following previous missing-modality learning works~\cite{first_prompt, dcp, syp}, we evaluate our method on three widely used vision-language benchmark datasets.

MM-IMDb~\cite{mmimdb} is a multimodal movie genre classification dataset containing 25,959 samples with paired movie posters and textual descriptions. Since each movie can belong to multiple genres, it is formulated as a multi-label classification task.

UPMC Food-101~\cite{food-101} contains noisy image-text pairs collected from the web across 101 food categories, which is used for multimodal food classification.

Hateful Memes~\cite{hatefulmemes} is a multimodal hateful content detection dataset consisting of over 10,000 image-text pairs, where both visual and textual information are required for accurate classification.

\begin{table*}[t]
\centering
\caption{
Comparison with the baseline on MM-IMDb~\cite{mmimdb}, UPMC Food-101~\cite{food-101}, and Hateful Memes~\cite{hatefulmemes} under different missing-modality conditions with missing rate $\eta=70\%$. ``Time (s)'' denotes the total adaptation time over the test set with one optimization step. Ours (1) and Ours (5) indicate TLP with 1 and 5 test-time optimization steps, respectively. The best results are highlighted in \textbf{bold}.
}
\label{tab:results_main}
\small 
% --- 关键：消除 booktabs 横线的额外间距，使竖线连续 ---
\setlength{\aboverulesep}{0pt}
\setlength{\belowrulesep}{0pt}
\renewcommand{\arraystretch}{1.0} % 稍微增加行高，弥补因消除间距导致的拥挤

\begin{tabular}{l|c|cc|cc| c| ccc} % 添加了竖线定义
\toprule % 使用标准 \hline 或 \toprule 均可，此时已不掉线
\multirow{2}{*}{Datasets} & \multirow{2}{*}{\begin{tabular}[c]{@{}c@{}}Missing\\ rate $\eta$\end{tabular}} & \multicolumn{2}{c|}{Train} & \multicolumn{2}{c|}{Test} & \multirow{2}{*}{Time (s)}   & \multirow{2}{*}{Baseline} & \multirow{2}{*}{Ours (1)} & \multirow{2}{*}{Ours (5)}\\  
` &  & Image & Text & Image & Text &  & \\ \hline
% 80.88
\multirow{3}{*}{\begin{tabular}[c]{@{}l@{}}MM-IMDb\\ (F1-Macro)\end{tabular}} 
 & \multirow{3}{*}{70\%} & 100\% & 30\% & 100\% & 30\% & 52.45 &44.76 & \cellcolor{gray!15}\textbf{46.91}   & \cellcolor{gray!15}\textbf{53.05} \\
 &  & 30\% & 100\% & 30\% & 100\% & 50.55 & 48.78 & \cellcolor{gray!15}\textbf{51.11} & \cellcolor{gray!15}\textbf{53.75} \\
 &  & 65\% & 65\%& 65\% & 65\% & 48.90 & 46.41  &\cellcolor{gray!15}\textbf{48.62}&\cellcolor{gray!15}\textbf{52.41} \\  \hline

\multirow{3}{*}{\begin{tabular}[c]{@{}l@{}}Food101\\ (Accuracy)\end{tabular}} 
 & \multirow{3}{*}{70\%} & 100\% & 30\% & 100\% & 30\% & 87.21 & 72.97 & \cellcolor{gray!15}\textbf{73.78}& \cellcolor{gray!15}\textbf{75.45} \\
 &  & 30\% & 100\%& 30\% & 100\% & 85.43 & 78.37 & \cellcolor{gray!15}\textbf{79.55} & \cellcolor{gray!15}\textbf{80.88} \\
 &  & 65\% & 65\% & 65\% & 65\% & 90.85  & 74.65 &\cellcolor{gray!15}\textbf{75.68} &\cellcolor{gray!15}\textbf{77.05} \\ 
 \midrule

\multirow{3}{*}{\begin{tabular}[c]{@{}l@{}}Hateful\\ Memes\\ (AUROC)\end{tabular}} 
 & \multirow{3}{*}{70\%} & 100\% & 30\% & 100\% & 30\% & 2.95 & 61.49  & \cellcolor{gray!15}\textbf{62.26} & \cellcolor{gray!15}\textbf{64.51} \\
 &  & 30\% & 100\% & 30\% & 100\% & 3.05 & 60.51  & \cellcolor{gray!15}\textbf{60.98} & \cellcolor{gray!15}\textbf{62.07} \\
 &  & 65\% & 65\%  & 65\% & 65\% & 2.81 & 61.52 & \cellcolor{gray!15}\textbf{61.80} & \cellcolor{gray!15}\textbf{62.61} \\
 \bottomrule
\end{tabular}
\end{table*}

\begin{figure*}[t]
\centering
\subfigure[Missing Text]{
    \includegraphics[width=0.3\textwidth, height=36mm]{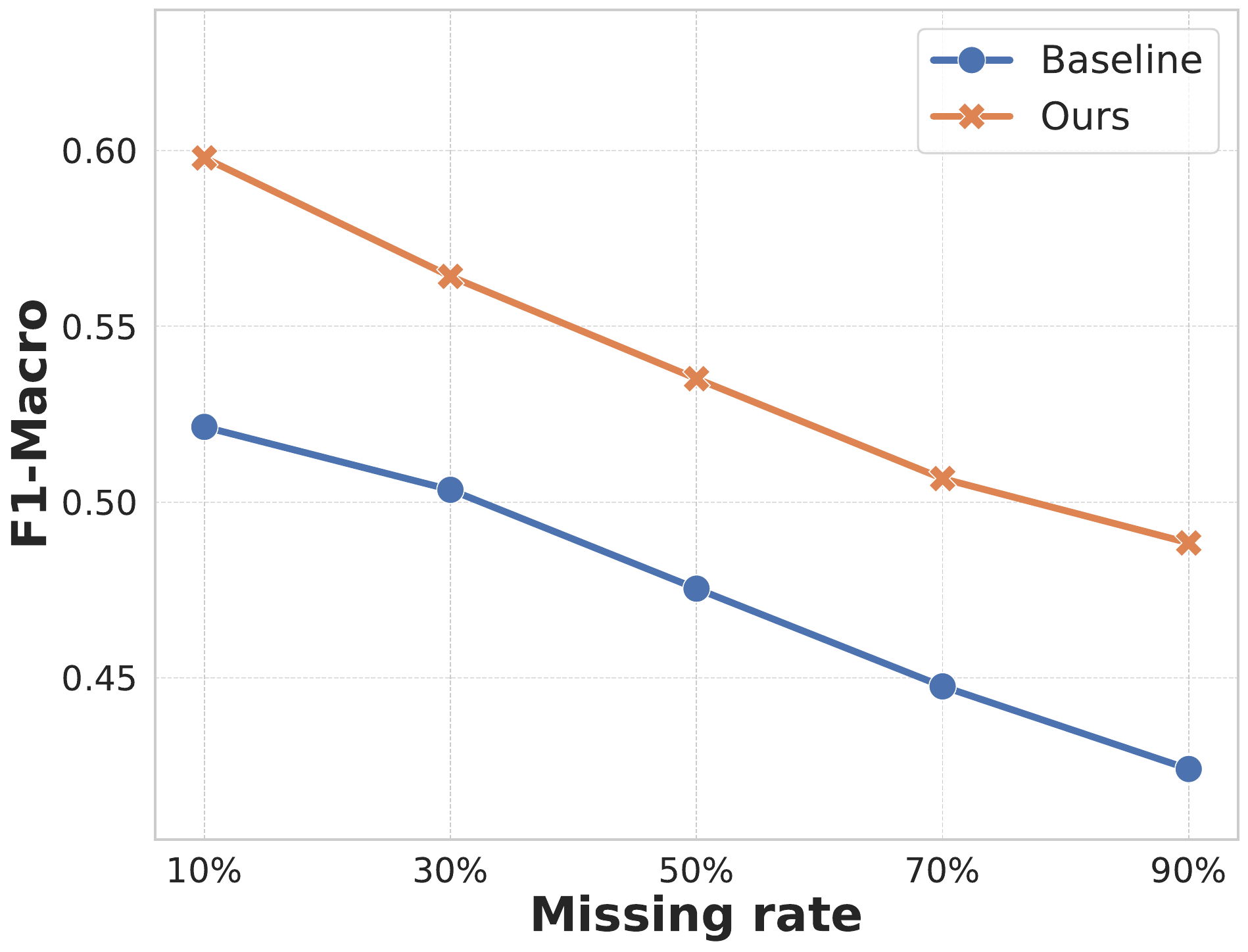}
}
\hfill
\subfigure[Missing Image]{
    \includegraphics[width=0.3\textwidth, height=36mm]{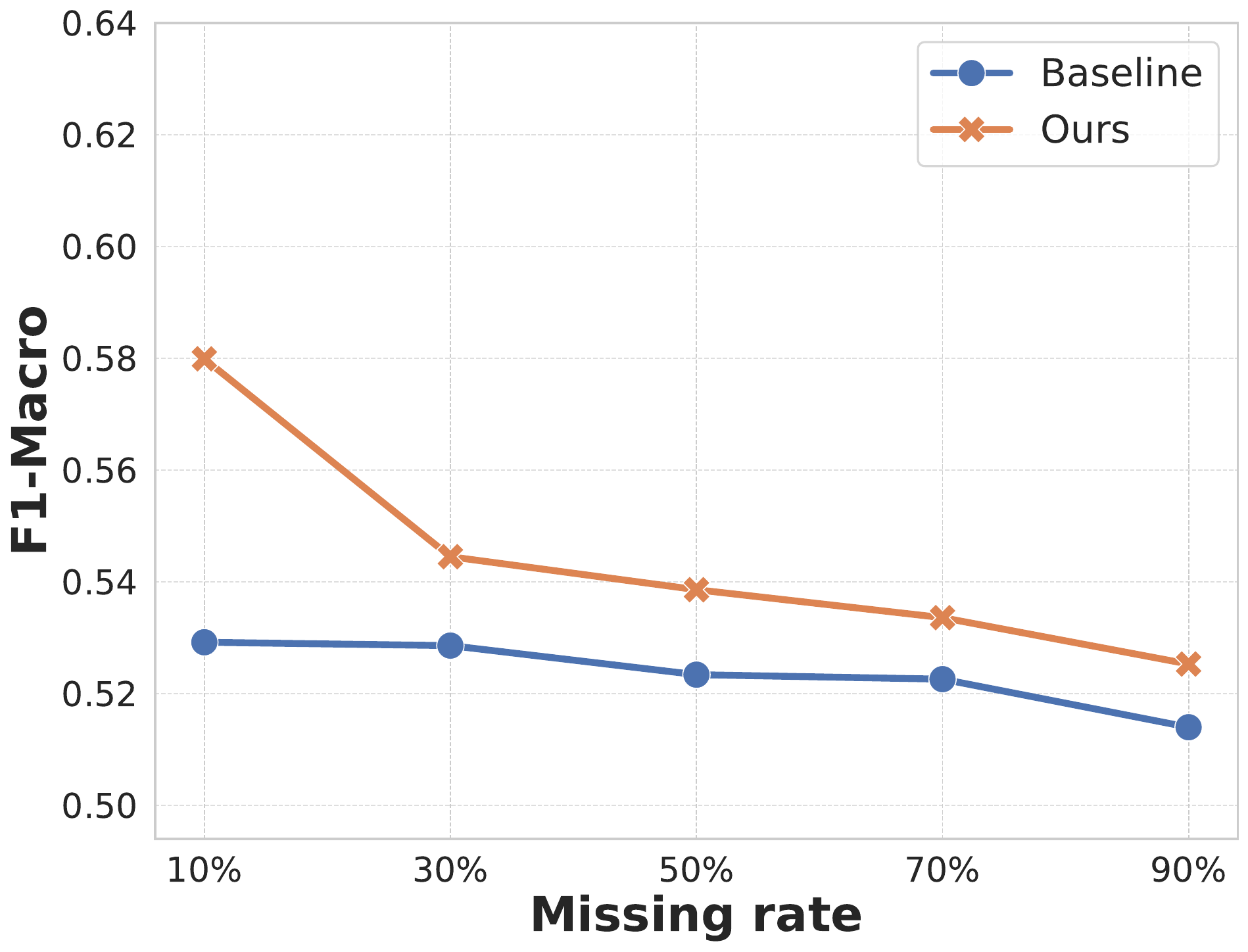}
}
\hfill
\subfigure[Missing Both]{

    \includegraphics[width=0.3\textwidth, height=36mm]{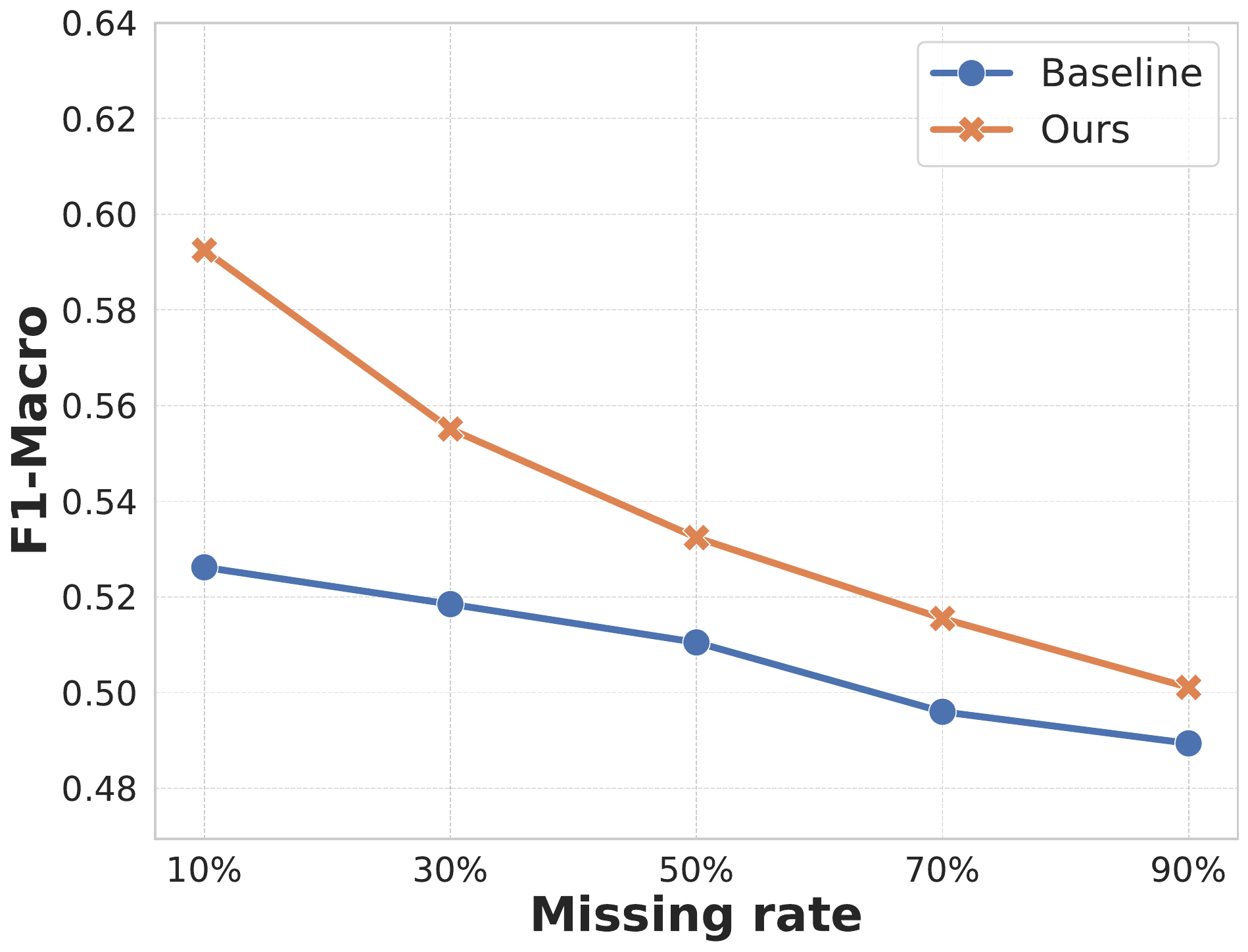}
    
}
    \caption{
Generalizability analysis of TLP under different test-time missing rates on the MM-IMDb dataset~\cite{mmimdb}. 
The source model is trained with complete modalities and evaluated under varying missing-modality ratios, where TLP is directly applied during inference without accessing source data. (a) Missing-text evaluation. (b) Missing-image evaluation. (c) Mixed missing evaluation with both text-only and image-only samples.
}
    % \end{subfigure}
    \label{fig:generlization}
\end{figure*}

\textbf{Implementation Details.}
We adopt CLIP~\cite{clip} with a ViT-B/16~\cite{vit} image encoder as the vision-language backbone. Input images are resized to $224\times224$, and text inputs are processed using the CLIP tokenizer with a maximum sequence length of 77. Following previous missing-modality settings~\cite{first_prompt,dcp,syp}, unavailable modalities are replaced with dummy inputs to maintain a consistent multimodal input format. During test-time adaptation, all parameters of the image encoder, text encoder, and task head remain frozen, and only the proposed logit prompt pool $\mathcal{P}$ is optimized. For each missing-modality condition, we introduce a class-level logit prompt with dimension $C$, resulting in only $3C$ trainable parameters. We optimize the logit prompts using the AdamW optimizer with a learning rate of $1\times10^{-2}$ and perform adaptation for $K$ optimization steps. For integration with training-based prompt learning methods, TLP is directly applied to their trained models without modifying the original training procedures. All experiments are conducted on a single NVIDIA RTX A6000 GPU with a batch size of 32.

\textbf{Missing Modality Setting.}
Following previous missing-modality learning works~\cite{first_prompt,dcp,syp}, we evaluate models under different modality-missing scenarios. The missing rate $\eta$ is defined as the proportion of samples with incomplete modalities. For vision-language tasks with image and text modalities, we consider three settings: missing-text, missing-image, and mixed missing. In the missing-text or missing-image setting, $\eta$ of samples contain only one available modality, while the remaining $(1-\eta)$ samples retain complete modalities. In the mixed missing setting, $\frac{\eta}{2}$ of samples are text-only, $\frac{\eta}{2}$ are image-only, and the remaining $(1-\eta)$ samples contain both modalities. This protocol can be extended to scenarios with $M$ modalities, where each incomplete case is sampled proportionally based on $\frac{\eta}{M^2-2}$, while $(1-\eta)$ of samples remain modality-complete.

\textbf{Baselines and Metrics.}
We evaluate TLP under two settings. First, we apply TLP to the frozen VLM backbone to validate its effectiveness for source-free test-time adaptation. Second, to evaluate its plug-and-play capability, we integrate TLP with representative training-based missing-modality prompt learning methods, including: (1) MMP~\cite{first_prompt}, which introduces missing-aware prompts to enhance multimodal representations; (2) DCP~\cite{dcp}, which models prompt correlations and cross-layer interactions for robust feature learning; and (3) SyP~\cite{syp}, which combines instance-aware features with learnable prompts through an adaptive prompting mechanism. Following prior works~\cite{first_prompt,dcp,syp}, we report F1-Macro for multi-label classification on MM-IMDb~\cite{mmimdb}, top-1 classification accuracy for recognition performance on UPMC Food-101~\cite{food-101}, and Area Under the Receiver Operating Characteristic Curve (AUROC) for hateful content detection on Hateful Memes~\cite{hatefulmemes}.

\begin{figure*}[t]
    \centering
    % 1. 放置全局图例图片 (关键步骤)
    % 假设你的图例文件名为 global_legend.pdf
    % 请确保该图片是横向展开的，且没有多余白边
     % \vspace{-9.8mm}
    \includegraphics[width=0.25\textwidth]{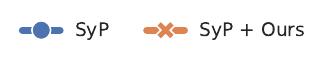}
    \vspace{-3.5mm}
    %[1.5ex] % 换行，并加一点垂直间距让它不拥挤

    % 2. 下方排列三个子图 (保持你原来的代码逻辑)
    % 稍微调整了宽度到 0.31\columnwidth 以填满整行
    \subfigure[Missing Text]{
        \includegraphics[width=0.3\textwidth, height=36mm]{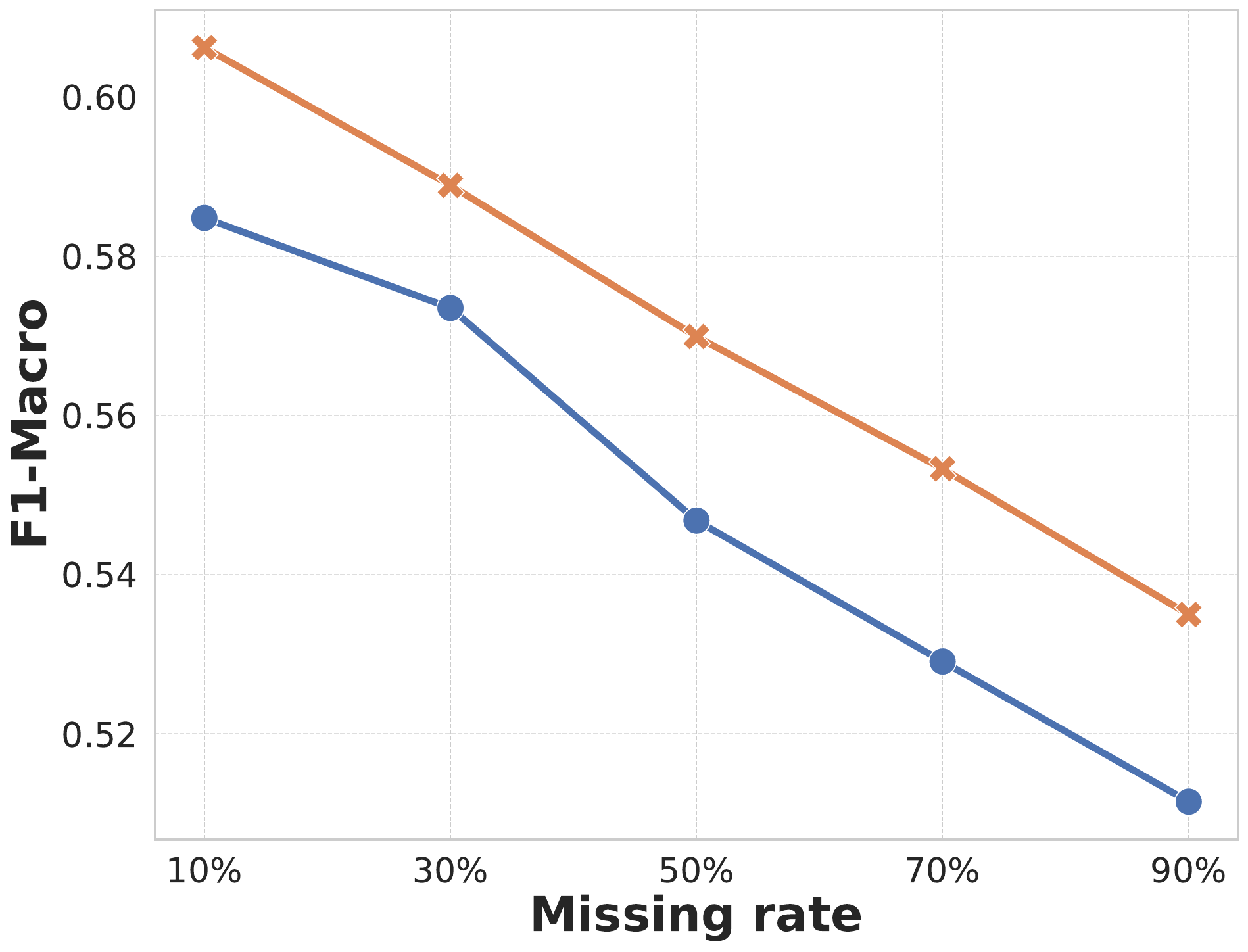}
    }
    \hfill % 在子图之间添加弹性间距，让它们均匀分布
    \subfigure[Missing Image]{
        \includegraphics[width=0.3\textwidth, height=36mm]{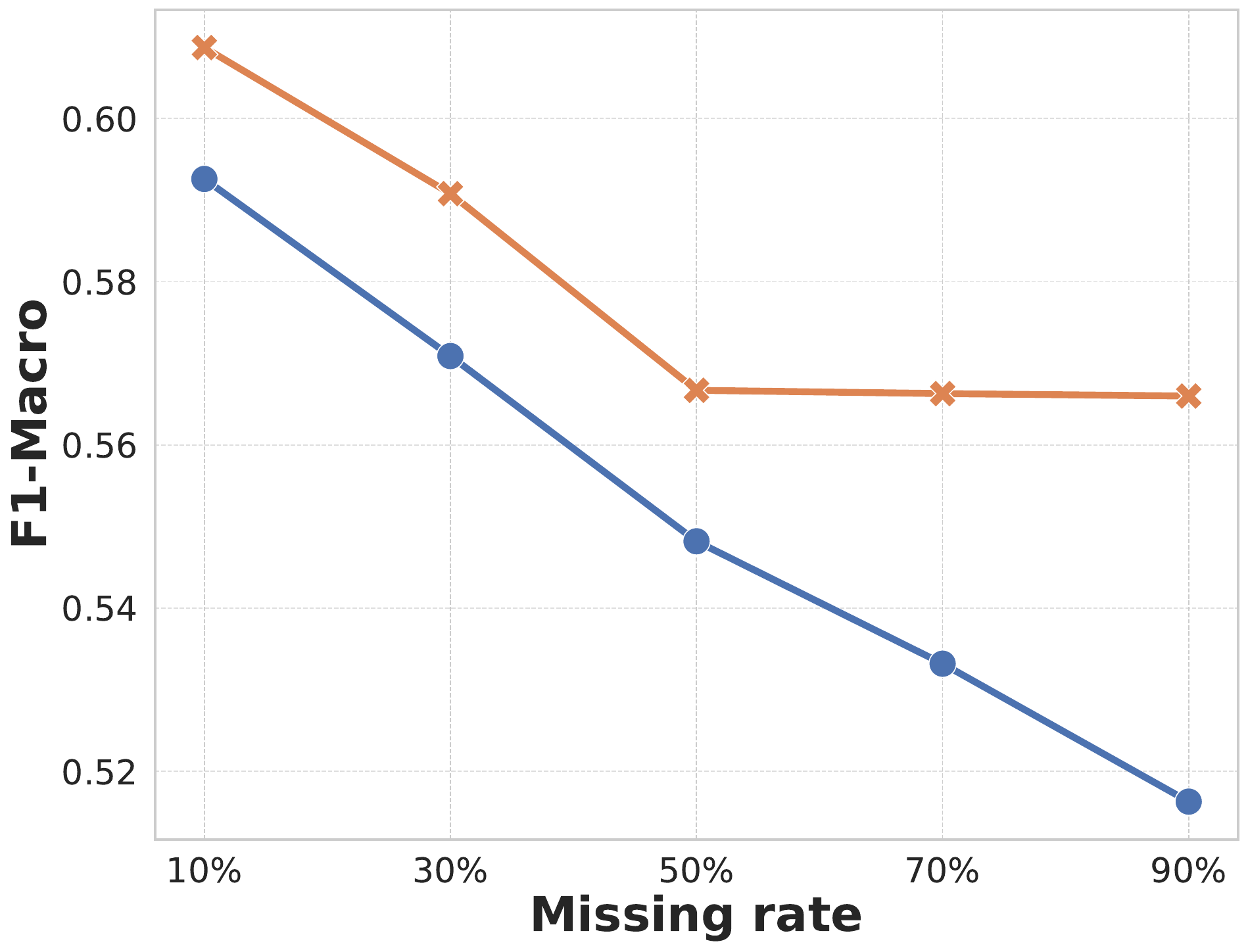}
    }
    \hfill
    \subfigure[Missing Both]{
        \includegraphics[width=0.3\textwidth, height=36mm]{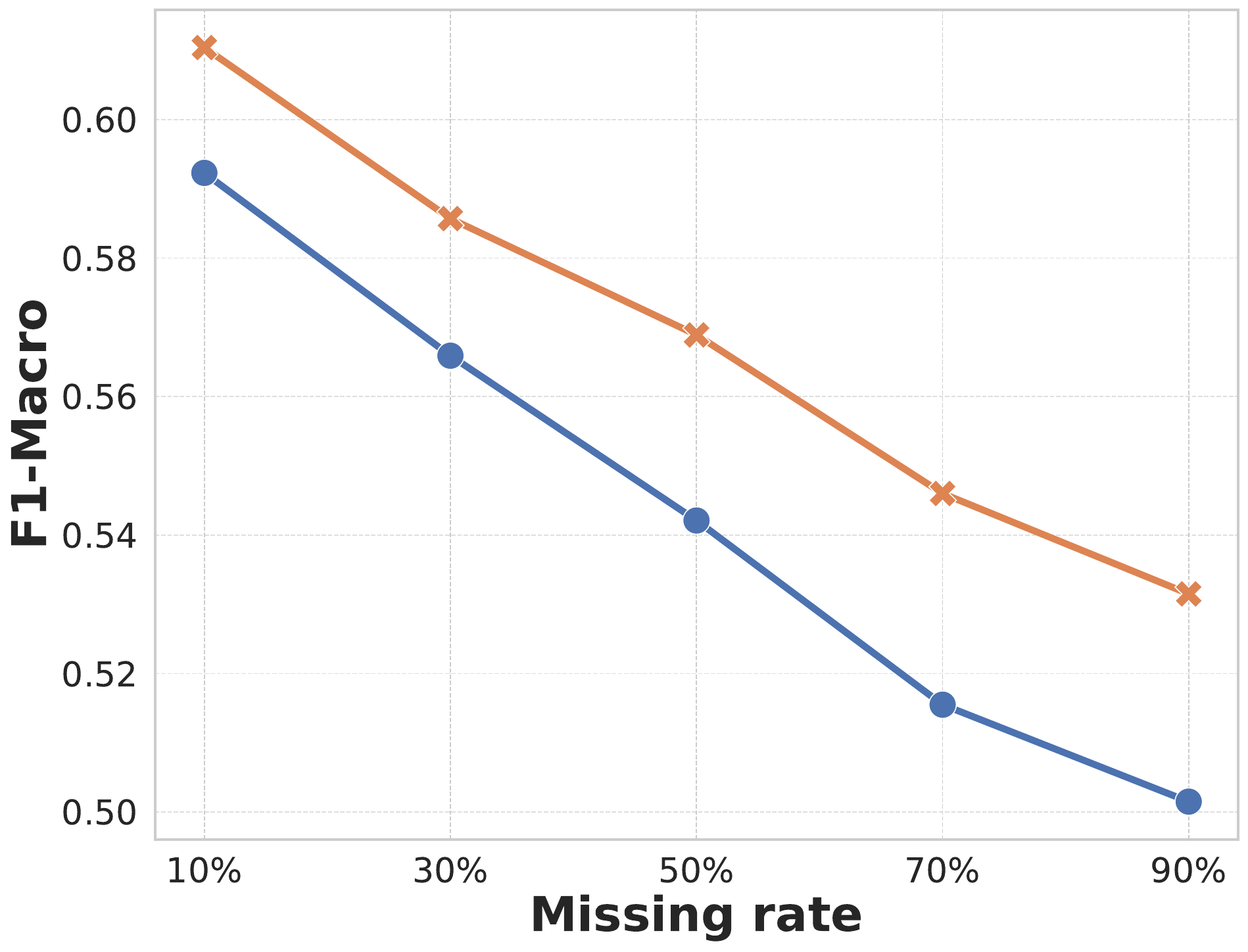}
    }

    \vspace{-2mm}
    \includegraphics[width=0.25\textwidth]{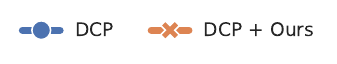}
    \vspace{-3.5mm}
    %[1.5ex] % 换行，并加一点垂直间距让它不拥挤

    % 2. 下方排列三个子图 (保持你原来的代码逻辑)
    % 稍微调整了宽度到 0.31\columnwidth 以填满整行
    \subfigure[Missing Text]{
        \includegraphics[width=0.3\textwidth, height=36mm]{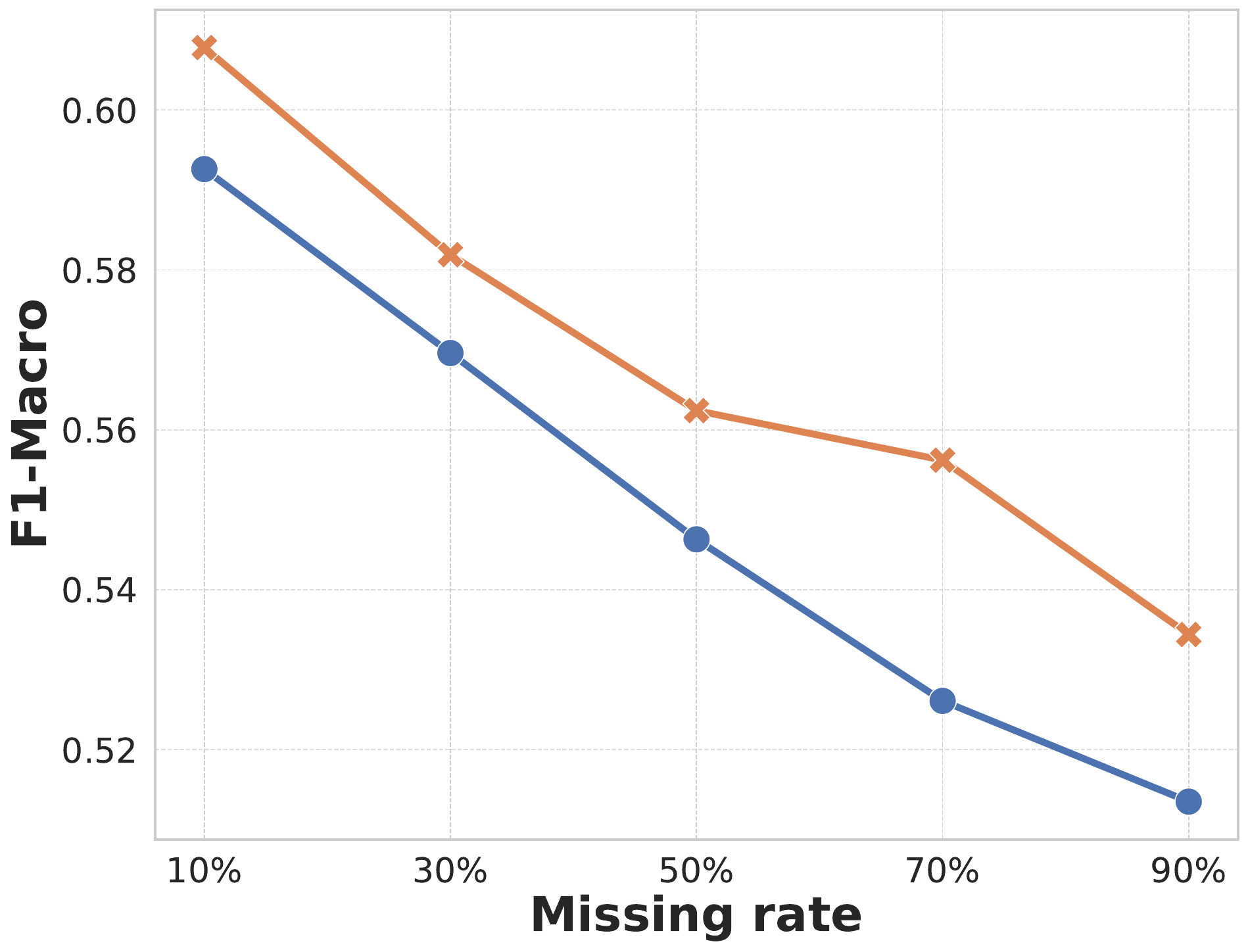}
    }
    \hfill % 在子图之间添加弹性间距，让它们均匀分布
    \subfigure[Missing Image]{
        \includegraphics[width=0.3\textwidth, height=36mm]{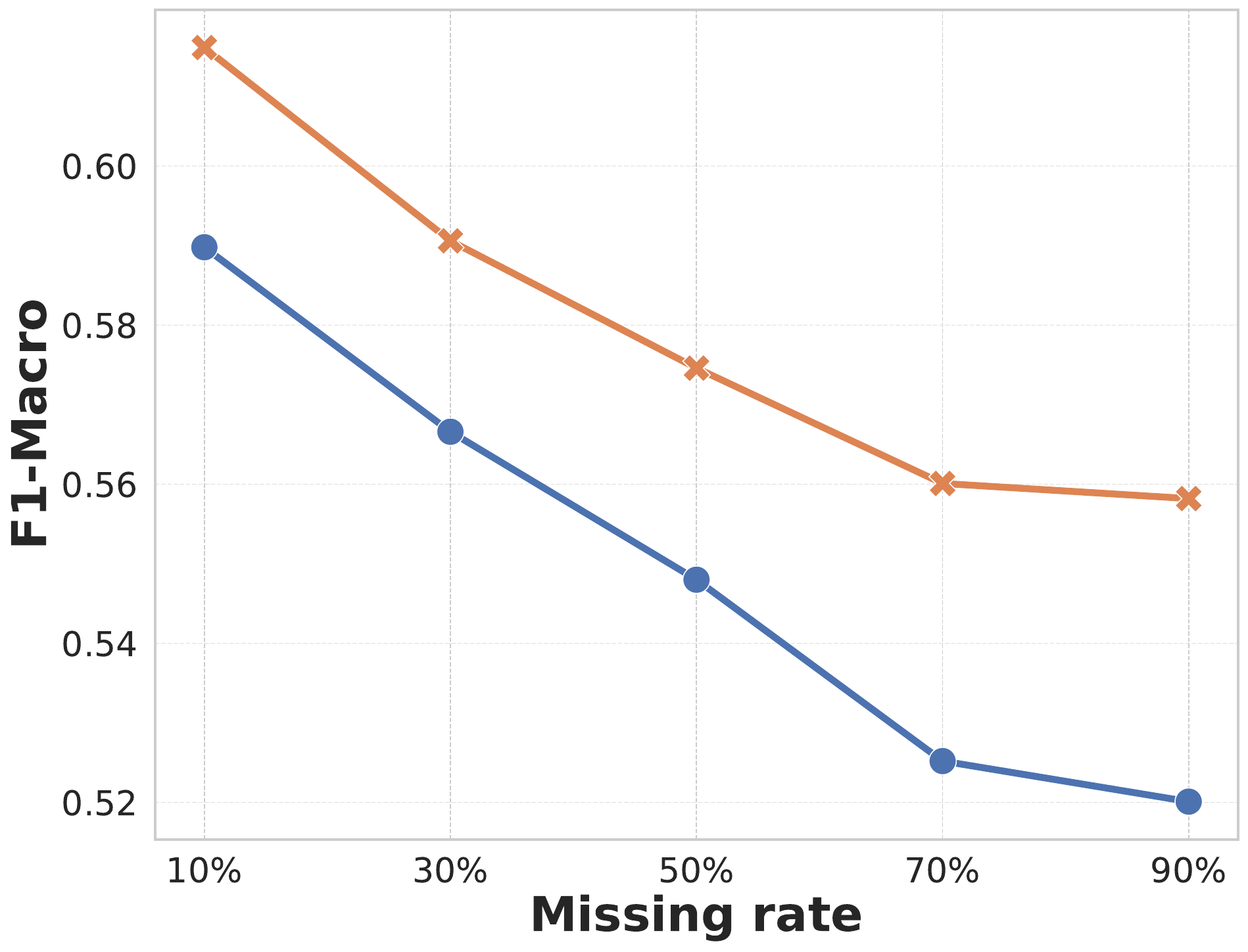}
    }
    \hfill
    \subfigure[Missing Both]{
        \includegraphics[width=0.3\textwidth, height=36mm]{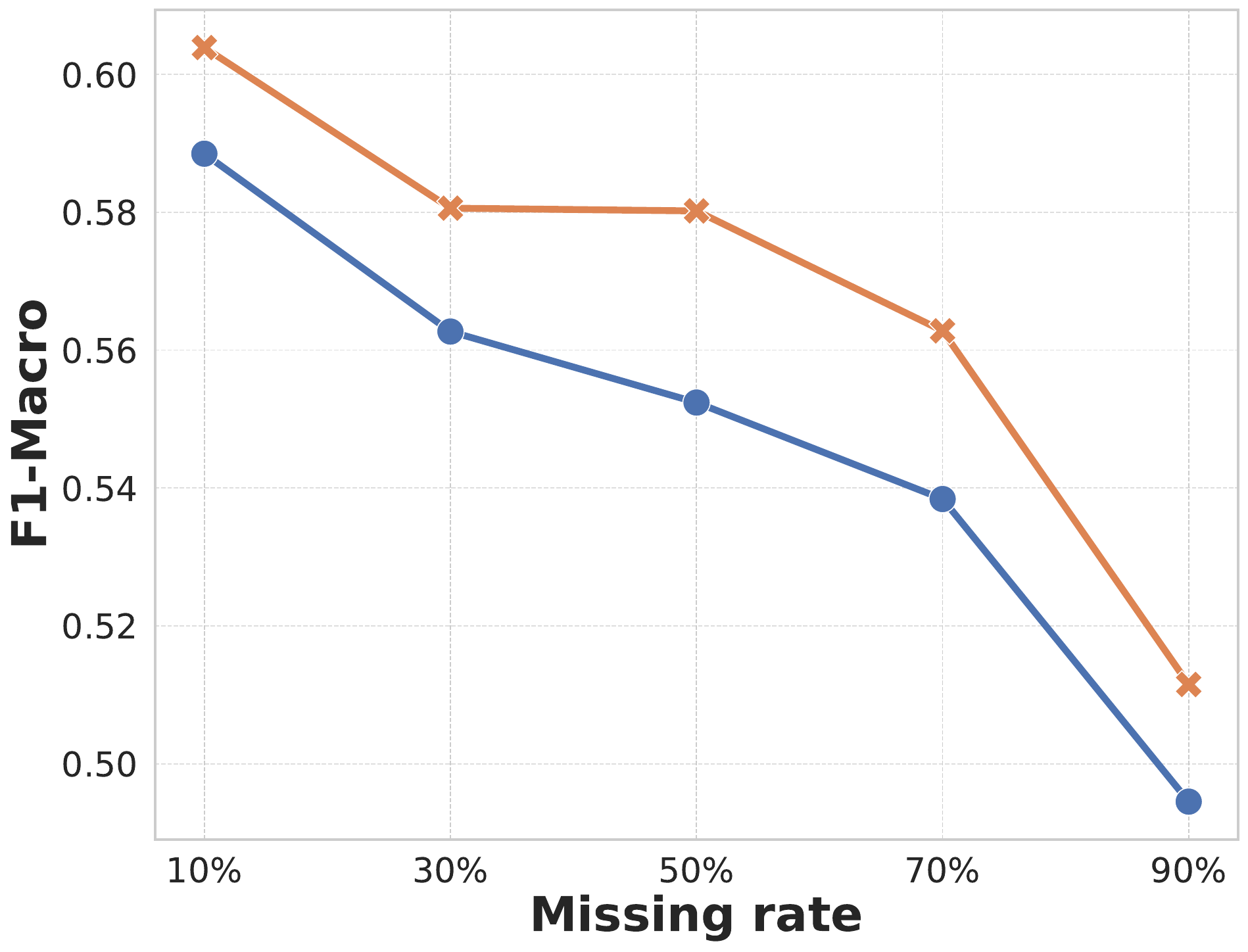}
    }

    \vspace{-2mm}
    \includegraphics[width=0.25\textwidth]{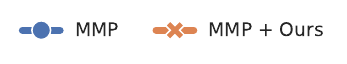}
    \vspace{-3.5mm}
    %[1.5ex] % 换行，并加一点垂直间距让它不拥挤

    % 2. 下方排列三个子图 (保持你原来的代码逻辑)
    % 稍微调整了宽度到 0.31\columnwidth 以填满整行
    \subfigure[Missing Text]{
        \includegraphics[width=0.3\textwidth, height=36mm]{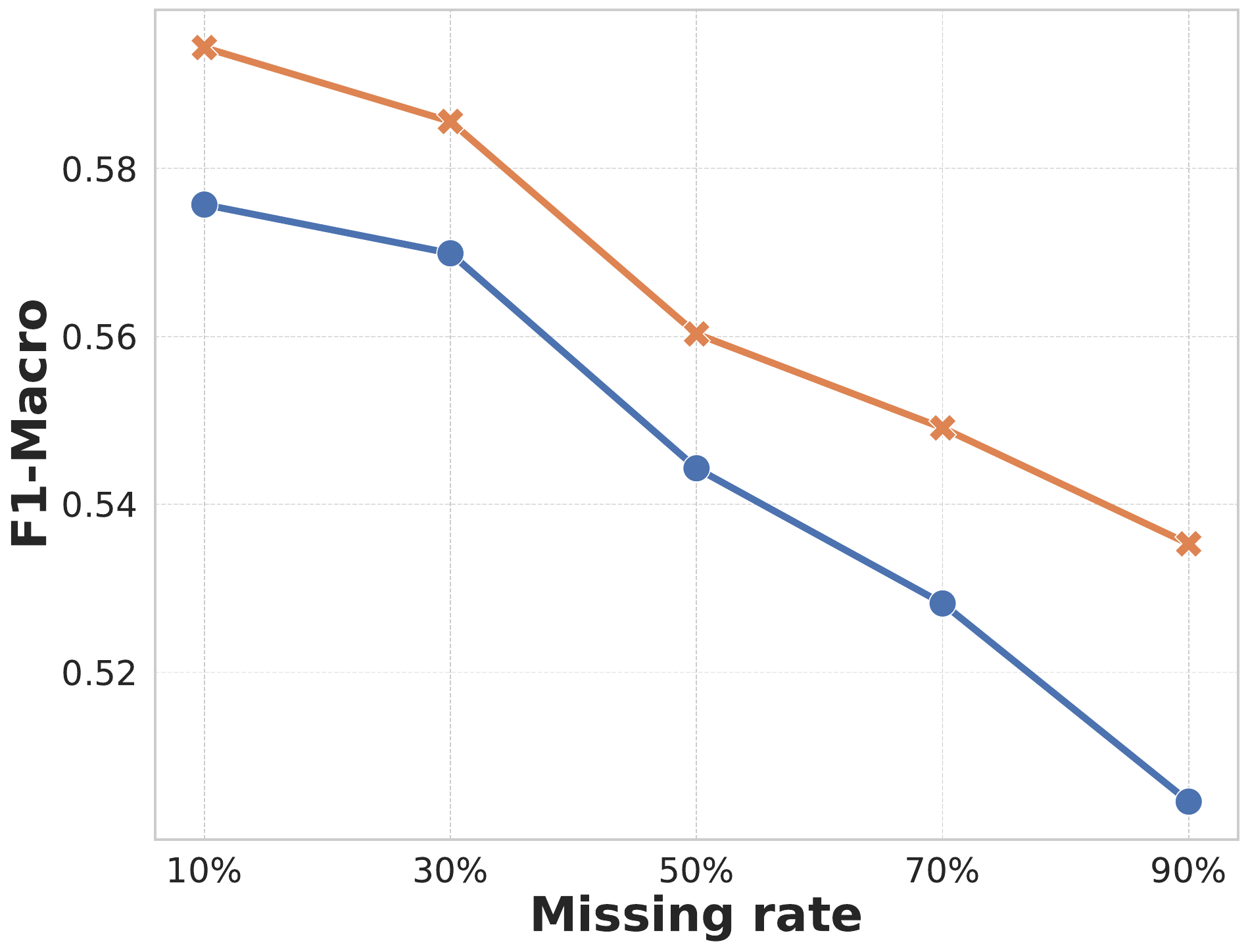}
    }
    \hfill % 在子图之间添加弹性间距，让它们均匀分布
    \subfigure[Missing Image]{
        \includegraphics[width=0.3\textwidth, height=36mm]{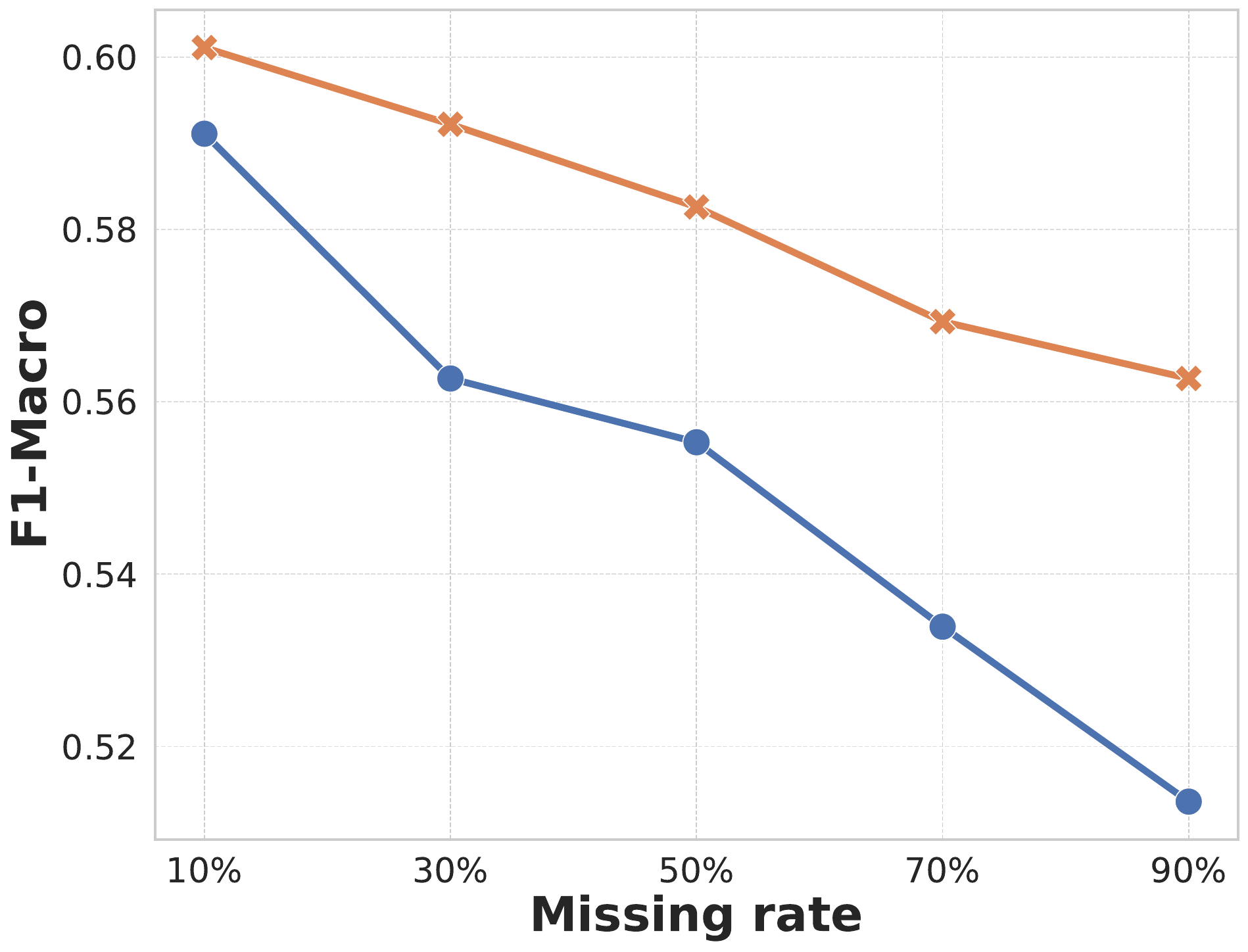}
    }
    \hfill
    \subfigure[Missing Both]{
        \includegraphics[width=0.3\textwidth, height=36mm]{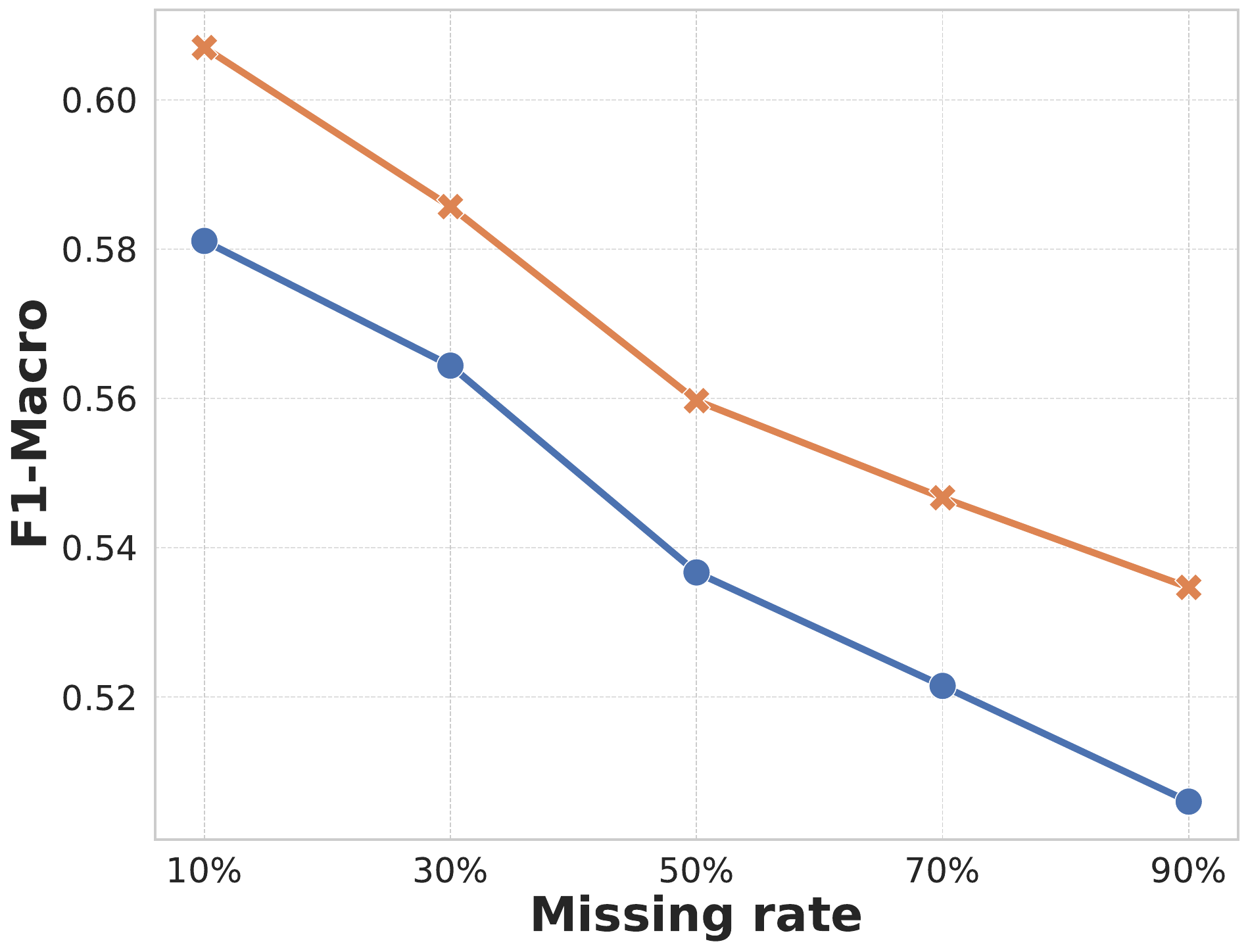}
    }
    
\caption{
Plug-and-play evaluation of TLP with existing missing-modality methods on MM-IMDb~\cite{mmimdb}. 
MMP~\cite{first_prompt}, DCP~\cite{dcp}, and SyP~\cite{syp} are trained and evaluated under corresponding missing-modality settings, while TLP is directly applied during inference without retraining.
}
    \label{fig:plug}
    % \vspace{-0.7cm}
\end{figure*}

\subsection{Experiment Results}

\textbf{Main results of comparison with baseline.} We first focus on studying the robustness of VLMs against modality incompleteness at test time without accessing source training data. The primary baseline is the source-trained VLM obtained under each missing-rate setting, which is directly evaluated on test samples with the same missing rate without any test-time adaptation. Since TLP keeps the entire backbone frozen and only optimizes lightweight logit prompts during inference, the performance improvement over this baseline directly reflects the benefit of source-free decision-space adaptation. As shown in Table~\ref{tab:results_main}, TLP consistently improves the performance of source-trained VLMs across different missing-modality scenarios without accessing source data. With only one test-time optimization step, TLP already brings clear improvements over the baseline (e.g., from 44.76\% to 46.91\%) while requiring only 3-90 seconds for adaptation. Increasing the optimization steps to five further enhances the performance, achieving improvements of up to 8\% compared with the non-adapted baseline. 

These results demonstrate that TLP can efficiently enhance the missing-modality robustness of VLMs through lightweight test-time adaptation. Notably, TLP achieves more substantial improvements on the challenging MM-IMDb dataset, where complex multi-label prediction requires effective utilization of incomplete multimodal information, further highlighting the benefit of decision-space adaptation.

\textbf{Generalization Analysis.}
To evaluate the generalization ability of TLP under changing missing-modality conditions, we conduct experiments where the source model is trained with complete modalities and evaluated under different test-time missing rates on the MM-IMDb dataset. As shown in Figure~\ref{fig:generlization}, TLP consistently outperforms the baseline across all missing scenarios, including missing text, missing image, and mixed missing settings, with missing rates ranging from 10\% to 90\%. 

The performance gains demonstrate that TLP can effectively adapt complete-trained VLMs to unseen missing-modality conditions without accessing source data or updating model parameters. Notably, missing text causes more severe performance degradation on MM-IMDb, indicating that textual information plays a critical role in this dataset. Even under this challenging scenario, TLP achieves substantial improvements over the baseline and maintains consistent gains as the missing rate increases. These results demonstrate that decision-space adaptation effectively improves the robustness of VLMs against varying degrees of modality incompleteness.

\textbf{Plug-and-play Evaluation.}
To further evaluate the compatibility of TLP with existing missing-modality learning methods, we integrate TLP with representative training-based prompt learning approaches, including MMP~\cite{first_prompt}, DCP~\cite{dcp}, and SyP~\cite{syp}. Since these methods also adopt prompting mechanisms for missing-modality adaptation, they provide suitable testbeds for evaluating the plug-and-play capability of TLP. These methods are first trained under corresponding missing-modality settings, and TLP is directly applied during inference without modifying their original training procedures.

As shown in Figure~\ref{fig:plug}, incorporating TLP consistently improves the performance of all compared methods across different missing scenarios and missing rates. This demonstrates that TLP can provide complementary benefits to existing training-based approaches by further adapting predictions at test time. Notably, the improvements remain stable and even become more pronounced as the missing rate increases, indicating that lightweight logit-level adaptation effectively enhances robustness under severe modality incompleteness. These results highlight the potential of TLP as a general plug-and-play test-time adaptation framework for missing-modality recognition.

\textbf{Ablation Analysis.}
We conduct ablation experiments to investigate the contribution of each optimization objective in TLP. As shown in Table~\ref{tab:ablation}, removing both objectives leads to inferior performance, demonstrating the necessity of source-free optimization for adapting logit prompts under missing-modality scenarios. Introducing either uncertainty-aware adjustment loss $\mathcal{L}_u$ or modality-complete consistency loss $\mathcal{L}_c$ improves the performance, validating the effectiveness of both components.

Specifically, $\mathcal{L}_u$ provides significant improvements on MM-IMDb~\cite{mmimdb} and Food101~\cite{food-101} by adjusting prediction uncertainty and alleviating biased confidence caused by incomplete modality information. Meanwhile, $\mathcal{L}_c$ effectively preserves semantic consistency by aligning missing-modality predictions with their nearest modality-complete references, leading to clear improvements especially on Hateful Memes~\cite{hatefulmemes}. By jointly optimizing these two objectives, TLP achieves the best performance across all benchmarks, demonstrating that uncertainty adjustment and semantic consistency provide complementary guidance for source-free logit prompt adaptation.

\textbf{Efficiency.}
We further analyze the efficiency of TLP in terms of trainable parameters and adaptation cost. Unlike existing prompt learning methods that optimize input-level prompts or additional modules, TLP keeps the entire VLM backbone frozen and only updates the logit prompt pool $\mathcal{P}$ during inference. Since each logit prompt contains only $C$ class-wise parameters, TLP introduces merely $3C$ learnable parameters for different missing-modality conditions. As shown in Table~\ref{tab:results_main}, TLP achieves consistent improvements with only a few test-time optimization steps. Even with a single adaptation step, TLP effectively improves recognition performance while requiring only limited adaptation time. These results demonstrate that TLP provides an efficient source-free adaptation solution for missing-modality recognition with minimal computational overhead. For example, TLP completes one-step adaptation within 3-90 seconds across different datasets while achieving noticeable gains.

\begin{figure}[t]
\centering
\subfigure[Missing Text]{
    \includegraphics[width=.2\textwidth]{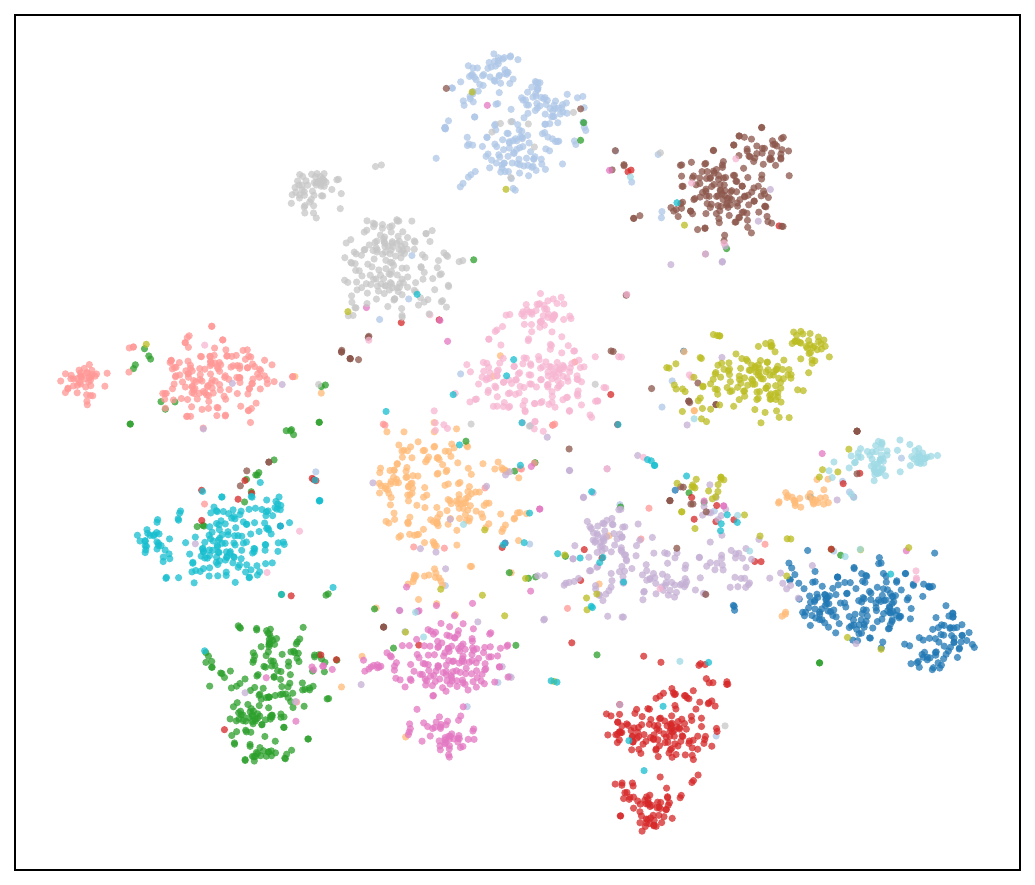}
}
\hfill
\subfigure[Missing Both]{
    \includegraphics[width=.2\textwidth]{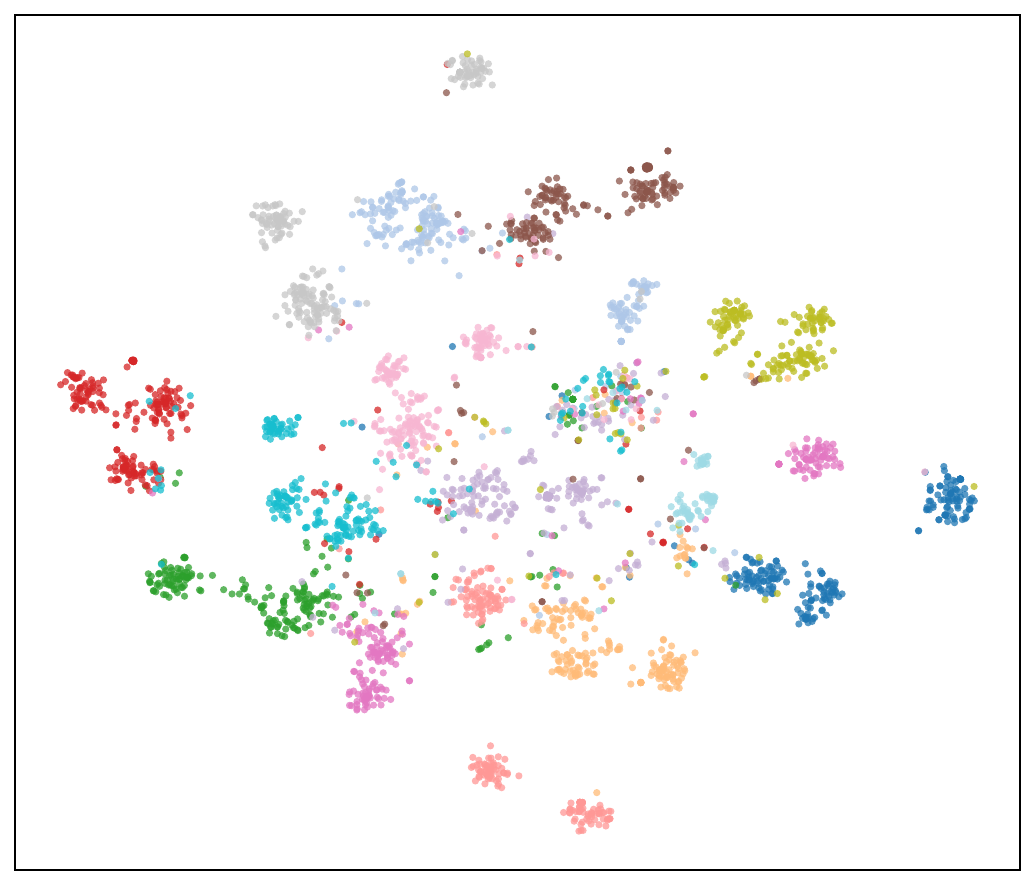}
}

% \hfill
% \subfigure[Missing Both]{

%     \includegraphics[width=0.3\textwidth, height=36mm]{ours_mmimdb_both.pdf}
    
% }
\caption{
Visualization analysis of TLP-adapted logit distributions on Food101~\cite{food-101} using t-SNE under different missing-modality scenarios.
}
    % \end{subfigure}
    \label{fig:vis}
\end{figure}

\textbf{Visualization Analysis.}
As shown in Figure~\ref{fig:vis}, we visualize the TLP-adapted logit distributions using t-SNE. Under different missing-modality scenarios, the adapted logits exhibit clear clustering structures in the decision space, indicating that TLP effectively preserves discriminative information through lightweight logit-level adaptation without updating the VLM backbone.

\section{Conclusion}

In this paper, we investigate source-free test-time adaptation for missing-modality recognition in vision-language models. We propose Test-Time Logit Prompting (TLP), a lightweight adaptation framework that improves frozen VLMs by optimizing missing-aware prompts directly in the logit space without accessing source training data. To achieve reliable prediction adjustment, TLP introduces uncertainty-aware adjustment and modality-complete consistency regularization, enabling effective adaptation under different missing-modality conditions while preserving semantic information. Extensive experiments on multiple vision-language benchmarks demonstrate that TLP consistently enhances missing-modality robustness with only a few optimization steps and hundreds of tunable parameters. Furthermore, TLP can be seamlessly integrated with existing training-based methods, highlighting its potential as an efficient plug-and-play framework for real-world incomplete multimodal scenarios.

\begin{table}[t]
\begin{tabular}[t]{cc|c c c}
\toprule
$L_u$ & $L_c$
& MM-IMDb
& Food101
& Hateful Memes\\
\midrule
$\checkmark$ & $\checkmark$ & \textbf{52.41} & \textbf{76.98} & \textbf{62.61}\\
$\checkmark$ &  & \underline{52.26} & \underline{76.77} & 61.55\\
 & $\checkmark$ & 46.74 & 75.03 & \underline{62.50}\\
 &&46.41&74.65&61.52\\
\bottomrule
\end{tabular}
\caption{
Ablation study of different optimization objectives in TLP on three benchmarks. \textbf{Bold} represents the best results.
}
\label{tab:ablation}
\end{table}

% \section{Conclusion}

% \newpage
\bigskip

\bibliography{aaai2027}

% Check whether the conference requires a reproducibility checklist to be included in the paper.
% If so, you can uncomment the following line and ajust the path to include it.
% \input{ReproducibilityChecklist.tex}

\end{document}